\documentclass[letterpaper,10pt,conference]{ieeeconf}
\IEEEoverridecommandlockouts
\usepackage[utf8]{inputenc}
\usepackage[T1]{fontenc}
\usepackage{graphicx,booktabs,multirow,makecell,array}
\usepackage[table]{xcolor}

\makeatletter
\newcommand{\resmergefullrow}[1]{%
  \rlap{\color{#1}\vrule width\columnwidth
    height\ht\@arstrutbox depth\dp\@arstrutbox}%
  \ignorespaces}
\makeatother
\usepackage{amsmath,amssymb,bm,mathtools}
\usepackage{booktabs}
\definecolor{oursblue}{RGB}{238,243,252}
\definecolor{bestblue}{RGB}{30,100,210}
\makeatletter
\let\NAT@parse\undefined
\makeatother
\usepackage[numbers,sort&compress]{natbib}

\usepackage[hidelinks]{hyperref}
\hypersetup{pdfauthor={Kefei Zhu, Lingyun Xu, Yue Xin, Xiaoqi Li, Caixin Guo, Xincheng Du, Yuanpei Chen, Yaodong Yang, Hao Dong},pdftitle={ResMerge: Residual Policy Learning and Merging for Continual Adaptation of Pre-trained Robot Policies},pdfsubject={},pdfkeywords={}}
\usepackage[capitalize,noabbrev]{cleveref}
\makeatletter
\let\resmerge@originalmakecaption\@makecaption
\long\def\@makecaption#1#2{%
  \ifx\@captype\@IEEEtablestring
    \begingroup
    \footnotesize\normalfont
    \setbox\@tempboxa\hbox{#1: #2}%
    \ifdim\wd\@tempboxa>\hsize
      \noindent #1: #2\par
    \else
      \hbox to\hsize{\hfil\box\@tempboxa\hfil}%
    \fi
    \endgroup
    \@IEEEtablecaptionsepspace
  \else
    \resmerge@originalmakecaption{#1}{#2}%
  \fi}
\makeatother
\title{ResMerge: Residual Policy Learning and Merging for Continual Adaptation of Pre-trained Robot Policies}
\author{Kefei Zhu$^{*}$, Lingyun Xu$^{*}$, Yue Xin$^{*}$, Xiaoqi Li$^{*}$, Caixin Guo,\\
Xincheng Du, Yuanpei Chen, Yaodong Yang, Hao Dong$^{\dagger}$\\[0.5ex]
{\small Peking University}
\thanks{$^{*}$Equal contribution. $^{\dagger}$Corresponding author.}}

\begin{document}

\IEEEaftertitletext{%
  \begin{center}
    \includegraphics[width=\textwidth]{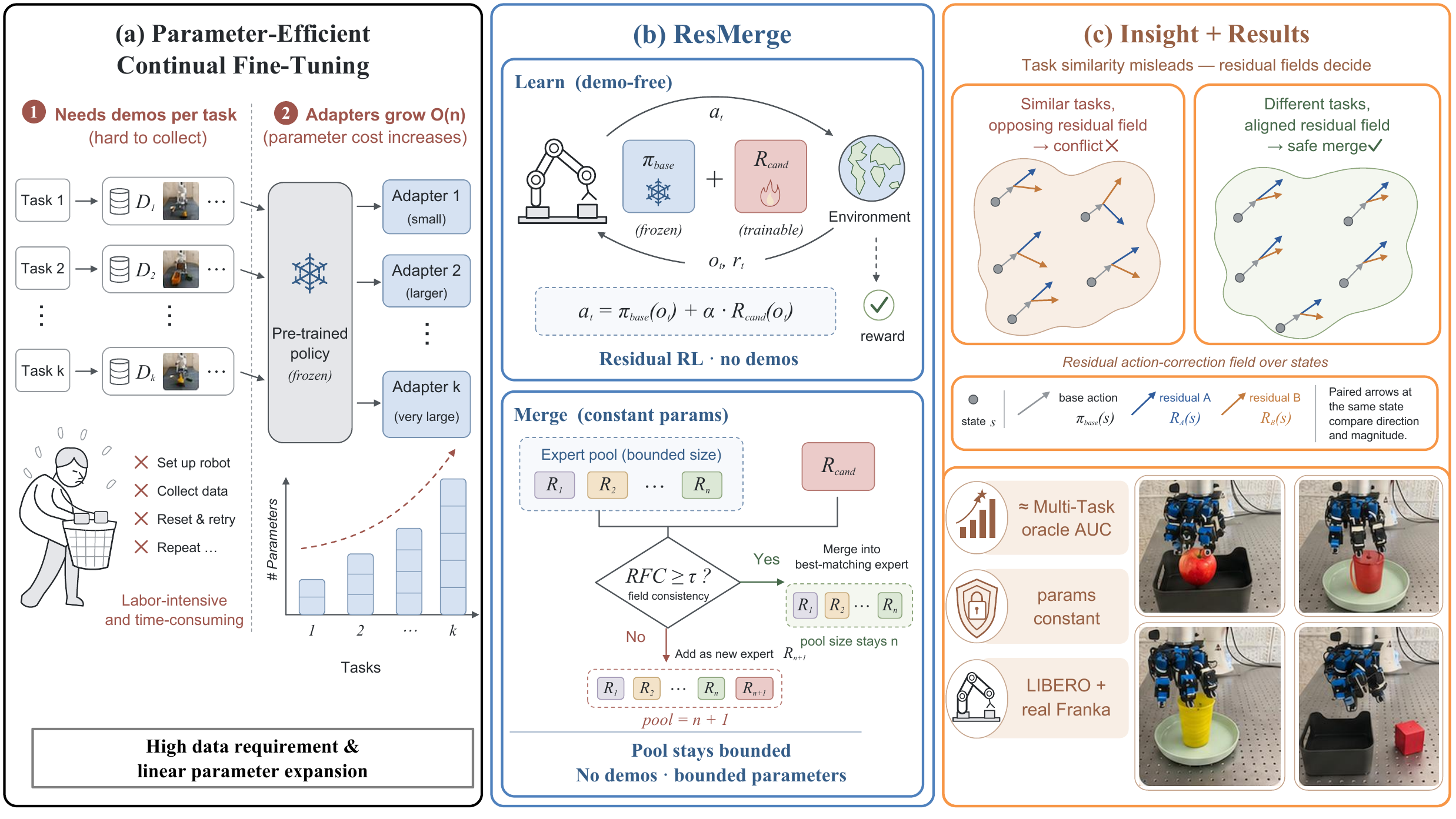}
    \vspace{2pt}
    \small
    \refstepcounter{figure}\label{fig:teaser_csq}
    \textbf{Fig. \thefigure. Continual adaptation with ResMerge.}
    \textbf{(a)} Adapter-per-task PECFT requires demonstrations and adds parameters for each task.
    \textbf{(b)} ResMerge learns residual corrections via RL without additional adaptation demonstrations,
    then merges compatible experts or expands the pool.
    \textbf{(c)} Task similarity can mislead; RFC measures consistency in direction and magnitude between residual fields.
  \end{center}
  \vspace{4pt}
}
\maketitle

\begin{abstract}
Continual adaptation to new tasks is essential for robots deployed in the real world, but adapting pre-trained policies remains challenging due to parameter growth and the need for continual demonstration collection. Parameter-Efficient Continual Fine-Tuning (PECFT) isolates task-specific updates, but adapter-per-task designs require demonstrations and grow with the task set. We propose \textbf{ResMerge}, a two-stage framework that learns \emph{mergeable} residual policies atop a frozen pre-trained policy: residual reinforcement learning (RL) acquires task-specific corrections without additional expert demonstrations during adaptation, and parameter-space merging consolidates compatible residuals to reduce stored parameters. We introduce \textit{Residual-Field Consistency (RFC)}, a simple, bounded, and interpretable metric that assesses compatibility in residual action-field space by comparing correction direction and magnitude. Experiments on LIBERO across two backbones and on a real-world Franka robot show improved continual learning performance over strong PECFT baselines, while selective merging reduces the residual-parameter footprint. Project website: \url{https://resmerge.github.io/}.
\end{abstract}

\section{Introduction}

For robots to remain useful in long-term deployment, they must continually adapt to new tasks while retaining previously acquired skills~\citep{lesort2020continual,lifelong,liu2023liberobenchmarkingknowledgetransfer}. Pre-trained robot policies provide useful manipulation priors~\citep{intelligence2025pi05visionlanguageactionmodelopenworld,o2024open}. Classical methods address forgetting through parameter isolation, expansion, or consolidation~\citep{mallya2018packnetaddingmultipletasks,rusu2016progressive,schwarz2018progress}. For large policies, \textit{Parameter-Efficient Continual Fine-Tuning (PECFT)} reduces adaptation costs~\citep{omaisan2025towards}, but sequential updates can still interfere~\citep{van2024continual}.

PECFT uses adapters~\citep{hu2022lora,kopiczko2023vera,liu2024tailtaskspecificadaptersimitation} or expert architectures~\citep{lei2025dynamic,wang2025self,qi2025adaptive} to isolate updates. Demonstration-trained, adapter-per-task designs require new demonstrations and accumulate parameters (Fig.~\ref{fig:teaser_csq}(a)). \textbf{ResMerge} (Fig.~\ref{fig:teaser_csq}(b)) uses sparse-reward residual reinforcement learning~\citep{johannink2019residual} to adapt without additional expert demonstrations, and merges compatible residuals in parameter space~\citep{yang2025modelmergingllmsmllms} to reduce storage.

\begin{figure*}[!t]
    \centering
    \includegraphics[width=\textwidth]{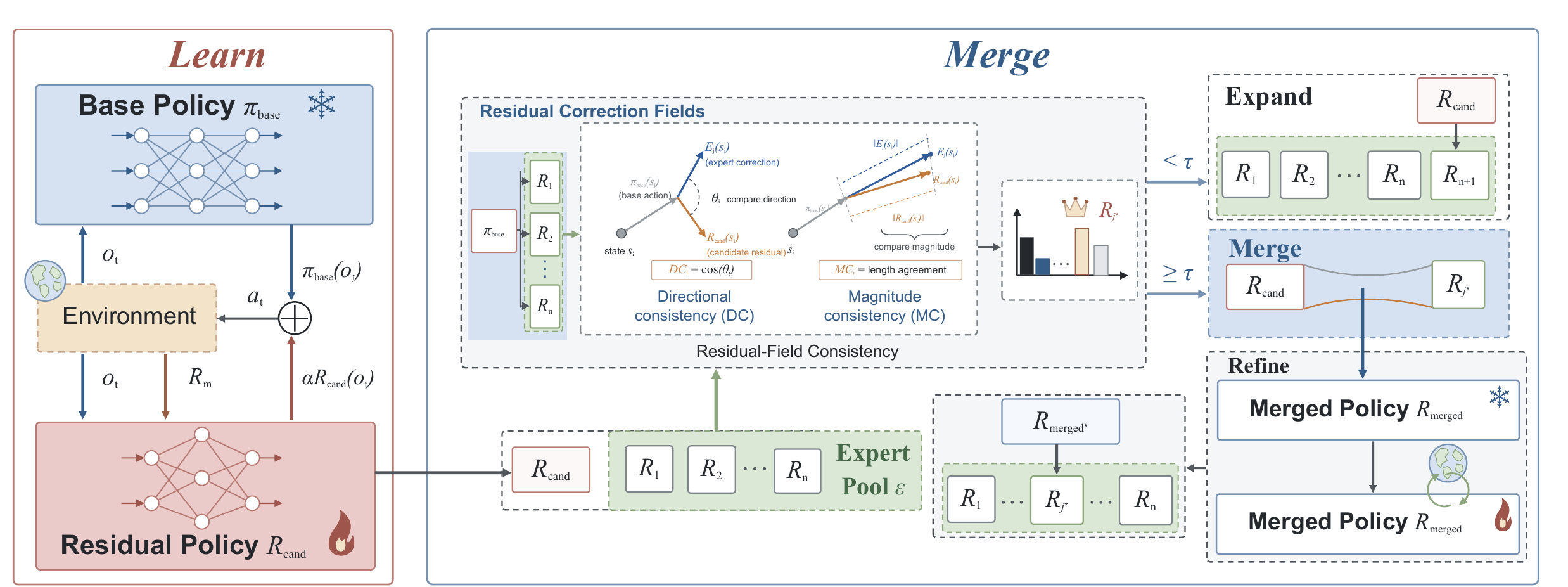}
\caption{\textbf{ResMerge framework.} In the \textbf{Learn} stage, a LoRA-parameterized residual policy $R_\text{cand}$ is trained on the new task via residual RL with $\pi_\text{base}$ frozen. In the \textbf{Merge} stage, $R_\text{cand}$ is merged into the most compatible existing expert when their \emph{Residual-Field Consistency (RFC)} meets or exceeds a compatibility threshold; otherwise it is added as a new expert.}
    \label{fig:method_overview}
    \end{figure*}

ResMerge has two stages (Fig.~\ref{fig:method_overview}). \textbf{Learn} trains a residual policy atop the frozen base policy via residual RL. We design its LoRA architecture for subsequent merging during parameter-efficient continual adaptation, extending residual learning beyond controller refinement. \textbf{Merge} combines compatible residual experts in parameter space, followed by RL refinement on the associated tasks.

A key problem is identifying residuals that preserve their corrections when merged. Task-level similarity~\citep{wan2024lotus,wang2022toward} is insufficient: similar tasks can require opposing corrections, while different tasks can share compatible corrections (Fig.~\ref{fig:teaser_csq}(c)). This identifies a gap in directly transplanting task-level merging criteria into residual RL. We therefore shift compatibility assessment from \emph{task space} to \emph{residual action-field space}. Our \textit{Residual-Field Consistency (RFC)} metric compares correction direction and magnitude on shared observations to select compatible experts for low-rank merging.

Our contributions are:
\begin{itemize}
    \item We propose \textbf{ResMerge} for continual adaptation without additional expert demonstrations. Its LoRA-parameterized residual architecture improves merging in our architecture comparison, reducing the parameters needed to retain learned skills.
    \item We introduce \textbf{Residual-Field Consistency (RFC)}, a \emph{simple, bounded, and interpretable} metric that shifts compatibility assessment from task-level similarity to agreement between residual action-correction fields.
    \item ResMerge consistently improves continual learning performance over strong PECFT baselines across two backbones on LIBERO and a real-world Franka robot, while selective merging reduces stored residual parameters.
\end{itemize}

\section{Related Work}
\label{sec:related}

\textbf{Continual Robot Learning.}
Continual robot learning balances skill acquisition and retention~\citep{lesort2020continual,liu2023liberobenchmarkingknowledgetransfer}, using successive demonstrations~\citep{yao2025think,xu2025speci,yang2024task,wan2024lotus,lee2024incremental,gu2024continual} or interaction with experience reuse~\citep{zhao2024experience,xie2022lifelong,mendez2023embodied,meng2025preserving,dong2025optimizing,malagon2025self}. For pre-trained policies, TAIL learns task-specific adapters~\citep{liu2024tailtaskspecificadaptersimitation}, while DMPEL combines low-rank experts, contextual routing, and coefficient replay~\citep{lei2025dynamic}. Separate modules increase storage. Progressive Neural Networks expand task-specific columns~\citep{rusu2016progressive}; Progress \& Compress consolidates knowledge through distillation~\citep{schwarz2018progress}.

Model merging consolidates capabilities in parameter space~\citep{yang2025modelmergingllmsmllms}. TIES-Merging addresses update interference~\citep{yadav2023ties}, AdaMerging learns coefficients~\citep{yang2023adamerging}, and RobustMerge considers low-rank direction robustness~\citep{zeng2025robustmerge}. Merging also supports robust VLA fine-tuning~\citep{yadav2025robust}. ResMerge consolidates compatible corrections to a shared controller without collecting additional expert demonstrations.

\textbf{Residual Reinforcement Learning.}
Residual reinforcement learning combines an existing controller with a learned action correction~\citep{johannink2019residual}. Residual Policy Learning~\citep{silver2018residualpolicy} improves hand-designed and model-predictive controllers with model-free RL, including in sparse-reward manipulation tasks. Policy Decorator~\citep{yuan2025policydecorator} extends this approach to large imitation policies and uses controlled exploration for online refinement. These methods establish residual learning as a way to improve a base controller through interaction while preserving its prior behavior.

ResMerge extends residual learning to support consolidation across a task sequence. We design a LoRA-parameterized actor with compact corrections that can be merged in parameter space. 

\section{Problem Setting}
\label{sec:problem-setting}
We study continual robot learning where a robot encounters a sequence of $K$ manipulation tasks $\{\mathcal{T}_1,\dots,\mathcal{T}_K\}$. Each task is a standard MDP with sparse reward.

\textbf{Pre-trained policy with continual adaptation.} The robot is initialized with a pre-trained base policy $\pi_{\text{base}}$ and observes the environment state $\bm{s}_t$ and a task description $l_k$ at each step. No task-specific demonstrations are provided. Tasks arrive sequentially: at task $k$, the agent should adapt to $\mathcal{T}_k$ while retaining performance on the previously seen tasks $\{\mathcal{T}_1,\dots,\mathcal{T}_{k-1}\}$.

\section{Method}
\label{sec:method}

\subsection{Learn Stage: Policy Learning}
\label{subsec:stage1}
In \textbf{Learn}, we train $R_{\text{cand}}$ to convergence on the current task while keeping the base policy frozen.

\textbf{Policy framework and observations.}
The base policy remains frozen and supplies a behavioral prior. The candidate residual learns additive action corrections:
\begin{equation}
 a_t=\pi_{\text{base}}(o_t^{\text{base}},l_k)+\alpha R_{\text{cand}}(o_t^{\text{res}}).
\end{equation}
Here $o_t^{\text{base}}$ and $o_t^{\text{res}}$ denote the inputs to the two policies, and $\alpha$ bounds the scale of the correction. In simulation, the base policy receives visual observations, proprioception, and a task instruction. The residual policy is \emph{state-based}: $o_t^{\text{res}}$ concatenates robot proprioception with simulator-provided object positions and orientations relative to the end effector. Proprioception includes joint positions and velocities, end-effector pose, and gripper state. This setup simplifies reinforcement learning and isolates the study of continual adaptation from representation learning. On the real robot, the residual instead receives frozen ViT image features and proprioception, without privileged simulator state.

\textbf{Residual policy learning.} We train the candidate with SAC~\citep{haarnoja2018softactorcriticoffpolicymaximum} under sparse task rewards while keeping the base policy frozen. The critic receives the residual observation and the final executed action. A warm-up period first exposes the agent to base-policy behavior; a Bernoulli mask with a progressively increasing activation probability then enables residual corrections during training. Evaluation fully enables the residual. We apply $\tanh$ squashing to the residual before scaling and use $\alpha=0.05$ in simulation.

\textbf{Low-rank residual actor.} The residual is a separate network, not an adapter inserted into the frozen backbone. Each hidden linear map is factorized as $y=BAx$, where $A\in\mathbb{R}^{r\times d_{\mathrm{in}}}$ and $B\in\mathbb{R}^{d_{\mathrm{out}}\times r}$. Three such layers, with LayerNorm and $\tanh$ activations, feed full-rank mean and log-standard-deviation heads. We refer to this factorized actor as the LoRA residual. We initialize the mean head to zero and the $B$ matrices with small Gaussian values ($\sigma=0.01$). Actor and critic encoders are separate: the full-rank critic learns task-specific value estimates without sharing encoder parameters with the low-rank actor. The critic is discarded after training; only the actor is retained for rollout and merging. This low-rank design keeps residuals compact and improves merging performance (Sec.~\ref{subsec:ablations}).

\subsection{Merge Stage: Residual Policy Merging}
\label{subsec:stage2}

\textbf{Why merging needs a residual-specific criterion.} Prior continual learning methods use task-level similarity~\citep{hiratani2024disentangling,wan2024lotus} to guide consolidation or routing. For residuals, compatibility instead depends on corrections to the shared controller: similar tasks may require opposing corrections, while different tasks may share compatible corrections. We therefore compare residual action fields directly.

\textbf{Residual-Field Consistency (RFC).}
A residual defines an action-correction field over its observation space. For a shared probe set of $M$ observations, write $u_i=R_{\text{cand}}(o_i^{\text{res}})$ and $v_i=R_j(o_i^{\text{res}})$ for the candidate and an existing expert. For this pair, we write $\mathrm{RFC}=\mathrm{RFC}(R_{\text{cand}},R_j)$ and define direction consistency (DC), magnitude consistency (MC), and RFC as
\begin{align}
\mathrm{DC}&=\frac{1}{M}\sum_{i=1}^{M}\frac{u_i^\top v_i}{\|u_i\|\,\|v_i\|+\epsilon},\label{eq:dc}\\
\mathrm{MC}&=\frac{1}{M}\sum_{i=1}^{M}\frac{2\min(\|u_i\|,\|v_i\|)}{\|u_i\|+\|v_i\|+\epsilon},\label{eq:mc}\\
\mathrm{RFC}&=\mathrm{DC}\cdot\mathrm{MC}\in[-1,1].\label{eq:rfc}
\end{align}
The constant $\epsilon>0$ stabilizes the ratios when correction magnitudes are near zero. DC measures average directional alignment over the shared observations; MC lies in $[0,1]$ and approaches one for matched magnitudes large relative to $\epsilon$. RFC combines these two aspects: large positive values indicate agreement in both direction and scale, while negative values indicate net directional opposition on the probe set. RFC guides expert selection; task performance also depends on merging and refinement.

\textbf{Merge decision and refinement.} We select $j^*=\arg\max_j\mathrm{RFC}(R_{\text{cand}},R_j)$ and merge into $R_{j^*}$ if its score is at least $\tau$; otherwise the candidate becomes a new expert. One threshold is fixed across tasks within a backbone. For OpenVLA-OFT and the real robot, $\tau=0.5$ is selected using held-out validation rollouts disjoint from test evaluation. The bank can still grow when no compatible expert is found.

When a merge is accepted, we combine the selected residuals layer-wise. Following robust model-merging operators~\citep{zeng2025robustmerge,yadav2025robust}, we prune low-magnitude rows of $BA$ from the smaller-magnitude source, normalize source scales on probe observations, and combine their $A$, $B$, and bias parameters by weighted averaging. LayerNorm parameters use the same convex weights. Rescaling each source's $B$ matrix controls scale differences before averaging. After merging, additional SAC updates on the affected tasks refine the consolidated actor. Reported post-merge success thus reflects both consolidation and refinement, rather than parameter averaging alone. The task-to-expert mapping is updated so that affected tasks select the merged expert.

\section{Experiments}

We evaluate ResMerge on LIBERO across two backbones and on four real-robot manipulation tasks. We ask: \textbf{(Q1)} Does ResMerge improve continual learning across backbone scales? \textbf{(Q2)} Can residual consolidation retain skills on a real robot? \textbf{(Q3)} How do RFC and the residual architecture affect merging?

\begin{figure*}[!t]
\centering
\includegraphics[width=\textwidth]{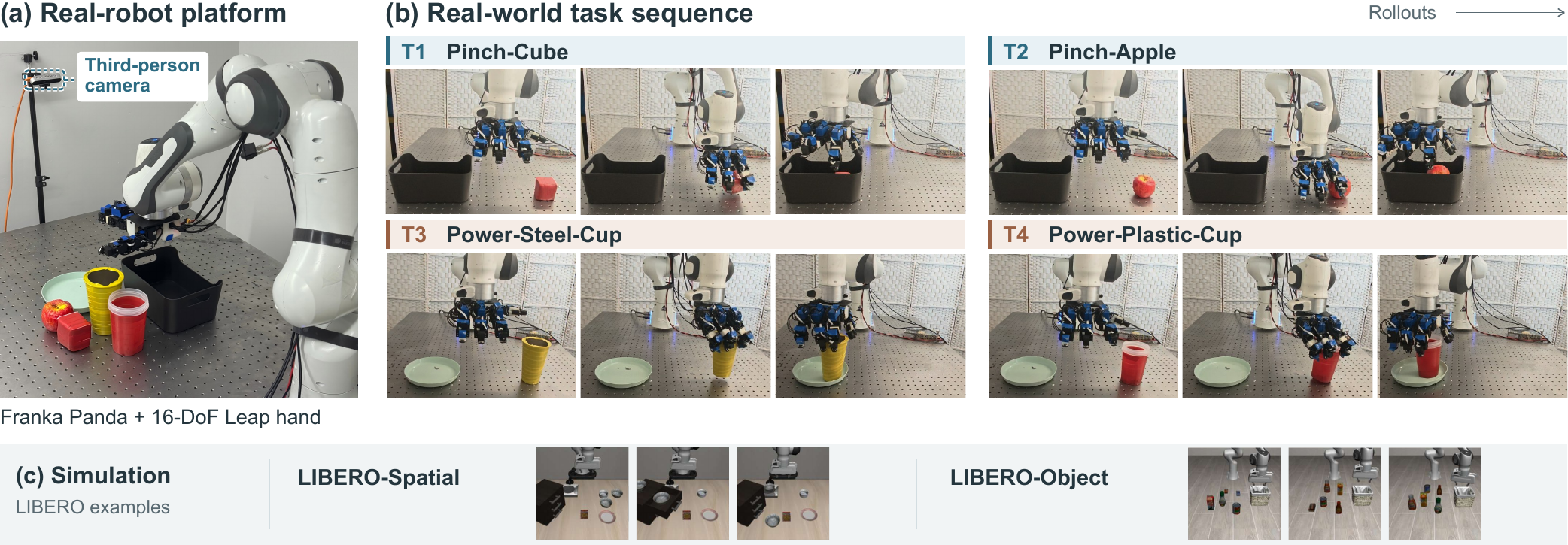}
\caption{\textbf{Overview of the experimental platforms and tasks.} A Franka Panda with a Leap hand performs four sequential pinch- and power-grasp tasks; three frames illustrate each rollout. The bottom strip shows representative LIBERO-Spatial and LIBERO-Object tasks used for simulation comparisons.}
\label{fig:experiment_overview}
\label{fig:realtasks}
\end{figure*}

\subsection{Experimental Setup}
\textbf{Tasks.} Our real-robot study uses a Franka Panda with a 16-DoF Leap hand to perform four sequential tabletop tasks (Fig.~\ref{fig:experiment_overview}). We additionally evaluate on LIBERO~\citep{liu2023liberobenchmarkingknowledgetransfer}, a continual robotic manipulation benchmark. We use LIBERO-Spatial and LIBERO-Object, each containing 10 tasks. ViT-T uses all ten tasks, while OpenVLA-OFT uses the five-task subsets specified below. The simulated platform is a Franka arm with a parallel gripper. All simulation experiments pair state-based residuals with a frozen visual base policy (Sec.~\ref{subsec:stage1}).

\textbf{Base policies and initialization.}\label{sec:base policy}
We use the 32M ViT-T BC policy from LIBERO~\citep{liu2023liberobenchmarkingknowledgetransfer} and the 8B OpenVLA-OFT VLA~\citep{kim2025finetuningvisionlanguageactionmodelsoptimizing} in simulation, and an IMLE BC policy~\citep{rana2025imlepolicy} on the real robot. ViT-T is pre-trained on LIBERO-90, then initialized with two demonstrations per target task. OpenVLA-OFT uses the released checkpoint from SimpleVLA-RL~\citep{li2025simplevlarlscalingvlatraining}, initialized by supervised fine-tuning on one trajectory from the target suite. These initialization demonstrations are outside the zero-additional-demonstration claim for residual adaptation. The real-robot base policy uses demonstrations from all four tasks. OpenVLA-OFT uses a learning rate of $3\times10^{-4}$ for both the residual actor and critic. 

ViT-T uses all ten tasks in each suite. For OpenVLA-OFT, we select five tasks per suite on which the initial base achieves nonzero success and use this same subset for all baselines. In order, LIBERO-Object task IDs are $(1,6,7,8,9)$ (cream cheese, butter, milk, chocolate pudding, orange juice into the basket); LIBERO-Spatial IDs are $(0,2,3,6,7)$ (black bowl between plate and ramekin, at table center, on the cookie box, next to the cookie box, or on the stove, moved onto the plate). This evaluates continual refinement of the base policy's existing skills.

\textbf{Baselines.} We compare ResMerge against (1) Sequential SFT, (2) Experience Replay~\citep{rolnick2019experiencereplaycontinuallearning}, (3) PackNet~\citep{mallya2018packnetaddingmultipletasks}, (4) TAIL~\citep{liu2024tailtaskspecificadaptersimitation}, (5) DMPEL~\citep{lei2025dynamic}, and a Multi-Task oracle trained jointly on all tasks (orange rows in Table~\ref{tab:main_results}). 

\textbf{ViT-T.} For ViT-T, we adopt the reported results from \citet{liu2023liberobenchmarkingknowledgetransfer} for four baseline methods: Sequential SFT, Experience Replay (ER), PackNet, and Multi-Task learning. These reported baselines use the same LIBERO suites; our residual-RL setup additionally uses state-based residual observations and online interaction. We implement TAIL and DMPEL according to their published methods and evaluate both on these same task suites.

\textbf{OpenVLA-OFT.} Seq.~SFT follows the OpenVLA-OFT recipe~\citep{kim2025finetuningvisionlanguageactionmodelsoptimizing}, with rank-32 LoRA on the vision-language backbone and trainable action and proprioception heads. It starts from the same initialized checkpoint (Sec.~\ref{sec:base policy}) and merges each task's adapter into the backbone before the next task. We use batch size 8 and learning rate $5\times10^{-4}$, stopping when L1 action loss reaches 0.05 and plateaus across consecutive epochs, instead of the original 0.01 criterion.

\textbf{Metrics.}\label{sec:metrics}
Let $N=K$ denote the number of tasks, $a_{i,j}$ the success rate on task $j$ after training on task $i$, and $P=N(N-1)/2$. We compute
\begin{align}
\mathrm{FWT}&=P^{-1}\sum_{i<j}a_{i,j},\\
\mathrm{NBT}&=P^{-1}\sum_{i>j}(a_{j,j}-a_{i,j}),\\
\mathrm{AUC}&=N^{-1}\sum_{i=1}^{N}\frac{1}{i}\sum_{j=1}^{i}a_{i,j}.
\end{align}
FWT here is average success on future tasks, without subtracting initial base success. NBT measures forgetting; negative values indicate improvement. AUC averages success on learned tasks across stages. Model Parameter Efficiency is $\mathrm{MPE}=\min(1,\sum_i|\Delta\theta_i^{\mathrm{ours}}|/\sum_i|\Delta\theta_i^{\mathrm{method}}|)$, where $\Delta\theta_i$ is added parameter overhead. MPE is normalized to ResMerge; $1.00$ does not imply a constant-size bank, which grows when candidates fail the RFC gate. Dashes in the MPE column denote non-comparable per-task footprints. The demonstration column counts additional expert demonstrations per task during continual adaptation, excluding base-policy initialization and online RL interaction. Success rates, FWT, and AUC are in percent; NBT is in percentage points (pp).

\begin{table*}[t]
\vspace*{6pt}
\caption{\textbf{Continual learning on LIBERO.} Results are mean $\pm$ standard deviation over five seeds. Demonstrations are counted only during adaptation.}
\label{tab:main_results}
\centering
\resizebox{\linewidth}{!}{
\begin{tabular}{ll|cc|ccc|ccc}
\toprule
\rowcolor{gray!15}
\textbf{Base Policy} & & & & \multicolumn{3}{c|}{\textbf{LIBERO-Spatial}} & \multicolumn{3}{c}{\textbf{LIBERO-Object}} \\
\rowcolor{gray!15}
\textbf{Architecture} & \raisebox{0.5\normalbaselineskip}[0pt][0pt]{\textbf{CL Method}} & \raisebox{0.5\normalbaselineskip}[0pt][0pt]{\textbf{MPE} $\uparrow$} & \raisebox{0.5\normalbaselineskip}[0pt][0pt]{\textbf{Demonstrations} $\downarrow$} & \textbf{FWT} $\uparrow$ & \textbf{NBT} $\downarrow$ & \textbf{AUC} $\uparrow$ & \textbf{FWT} $\uparrow$ & \textbf{NBT} $\downarrow$ & \textbf{AUC} $\uparrow$ \\
\midrule
\multirow{8}{*}{\textbf{ViT-T}}
 & Seq. SFT & --- & 50 demos/task & $63 \pm 2$ & $76 \pm 1$ & $16 \pm 1$ & $\mathbf{76 \pm 3}$ & $73 \pm 3$ & $27 \pm 2$ \\
 & ER & --- & 50 demos/task & $\mathbf{63 \pm 1}$ & $29 \pm 2$ & $50 \pm 2$ & $70 \pm 2$ & $28 \pm 1$ & $57 \pm 1$ \\
 & PackNet & --- & 50 demos/task & $57 \pm 4$ & $15 \pm 1$ & $59 \pm 3$ & $58 \pm 3$ & $18 \pm 2$ & $56 \pm 4$ \\
 & TAIL & $0.70$ & 50 demos/task & $51 \pm 2$ & $56 \pm 1$ & $60 \pm 3$ & $58 \pm 2$ & $50 \pm 2$ & $70 \pm 1$ \\
 & DMPEL & $0.69$ & 50 demos/task & $22 \pm 2$ & $28 \pm 3$ & $29 \pm 2$ & $14 \pm 3$ & $25 \pm 1$ & $27 \pm 3$ \\
\cmidrule{2-10}
\rowcolor{blue!8}
 & \textbf{ResMerge (w/o merge)} & $0.69$ & \textcolor{blue!70!black}{\textbf{0 demos/task}} & $31 \pm 3$ & $10 \pm 2$ & $43 \pm 1$ & $27 \pm 2$ & $13 \pm 3$ & $54 \pm 2$ \\
\rowcolor{blue!8}
 & \textbf{ResMerge (ours)} & \textcolor{blue!70!black}{$\mathbf{1.00}$} & \textcolor{blue!70!black}{\textbf{0 demos/task}} & $42 \pm 1$ & \textcolor{red!60!pink}{$\mathbf{-3 \pm 1}$} & \textcolor{red!60!pink}{$\mathbf{83 \pm 3}$} & $49 \pm 2$ & \textcolor{red!60!pink}{$\mathbf{-8 \pm 1}$} & \textcolor{red!60!pink}{$\mathbf{79 \pm 1}$} \\
\cmidrule{2-10}
\rowcolor{orange!12}
 & Multi-Task (oracle) & --- & 50 demos/task & --- & --- & $79 \pm 1$ & --- & --- & $78 \pm 2$ \\
\midrule
\multirow{4}{*}[3ex]{\textbf{OpenVLA-OFT}}
 & Seq. SFT (LoRA) & --- & 50 demos/task & $0 \pm 0$ & $98 \pm 2$ & $47 \pm 1$ & $10 \pm 1$ & $70 \pm 4$ & $62 \pm 1$ \\
\cmidrule{2-10}
\rowcolor{blue!8}
 & \textbf{ResMerge (w/o merge)} & $0.80$ & \textcolor{blue!70!black}{\textbf{0 demos/task}} & $68\pm 1 $ & \textcolor{blue!70!black}{$\mathbf{-3 \pm 1}$} & $81 \pm 1$ & \textcolor{blue!70!black}{$\mathbf{58\pm 2}$} & $2 \pm 1 $ & \textcolor{blue!70!black}{$\mathbf{90 \pm 2}$}\\
\rowcolor{blue!8}
 & \textbf{ResMerge (ours)} & \textcolor{blue!70!black}{$\mathbf{1.00}$} & \textcolor{blue!70!black}{\textbf{0 demos/task}} & \textcolor{blue!70!black}{$\mathbf{77 \pm 3}$} & $2 \pm 1$ & \textcolor{blue!70!black}{$\mathbf{95 \pm 2}$} & $53 \pm 1 $ & \textcolor{blue!70!black}{$\mathbf{1 \pm 1}$} & $87 \pm 1 $ \\
\cmidrule{2-10}
\rowcolor{orange!12}
 & Multi-Task (oracle) & --- & 50 demos/task & --- & --- & $95 \pm 1$ & --- & --- & $94 \pm 1$ \\
\bottomrule
\end{tabular}
}
\vspace*{-6pt}
\end{table*}

\subsection{Simulation Results}
\textbf{Overall comparison.}
Table~\ref{tab:main_results} compares two backbones on both LIBERO suites: \textcolor{blue!70!black}{blue} highlights leading ResMerge results, and \textcolor{red!60!pink}{red} highlights ViT-T retention and AUC. On the larger OpenVLA-OFT backbone, ResMerge achieves favorable transfer metrics: FWT \textcolor{blue!70!black}{77\%}/53\% (Spatial/Object), NBT 2/\textcolor{blue!70!black}{1} pp, and AUC \textcolor{blue!70!black}{95\%}/87\%, approaching the Multi-Task oracle (95\%/94\% AUC) without additional expert demonstrations during residual adaptation. The baseline at this scale, Seq.~SFT (LoRA), still suffers substantial forgetting (NBT 98/70 pp) and trails ResMerge by 48/25 AUC points (AUC 47\%/62\%), while using 50 demonstrations per task. On ViT-T, ResMerge attains the highest AUC (\textcolor{red!60!pink}{83\%/79\%}) and lowest NBT (\textcolor{red!60!pink}{$-3$/$-8$ pp}), indicating positive backward transfer. Both variants use \textcolor{blue!70!black}{0 additional expert demonstrations per task} during adaptation.

\textbf{Plasticity-stability trade-off.} Some baselines (Seq.~SFT, ER) reach higher single-suite FWT on ViT-T but incur large NBT and therefore low AUC---the classical plasticity-stability trade-off~\citep{van2024continual}. We primarily evaluate AUC, which reflects both learning and retention. Consolidation also involves a trade-off: on OpenVLA-OFT/Object, no-merge reaches AUC \textcolor{blue!70!black}{90\%} and FWT \textcolor{blue!70!black}{58\%}, compared with 87\% and 53\% for ResMerge. Thus merging does not improve every metric on every suite.

\begin{figure*}[t]
    \centering
    \includegraphics[width=\linewidth]{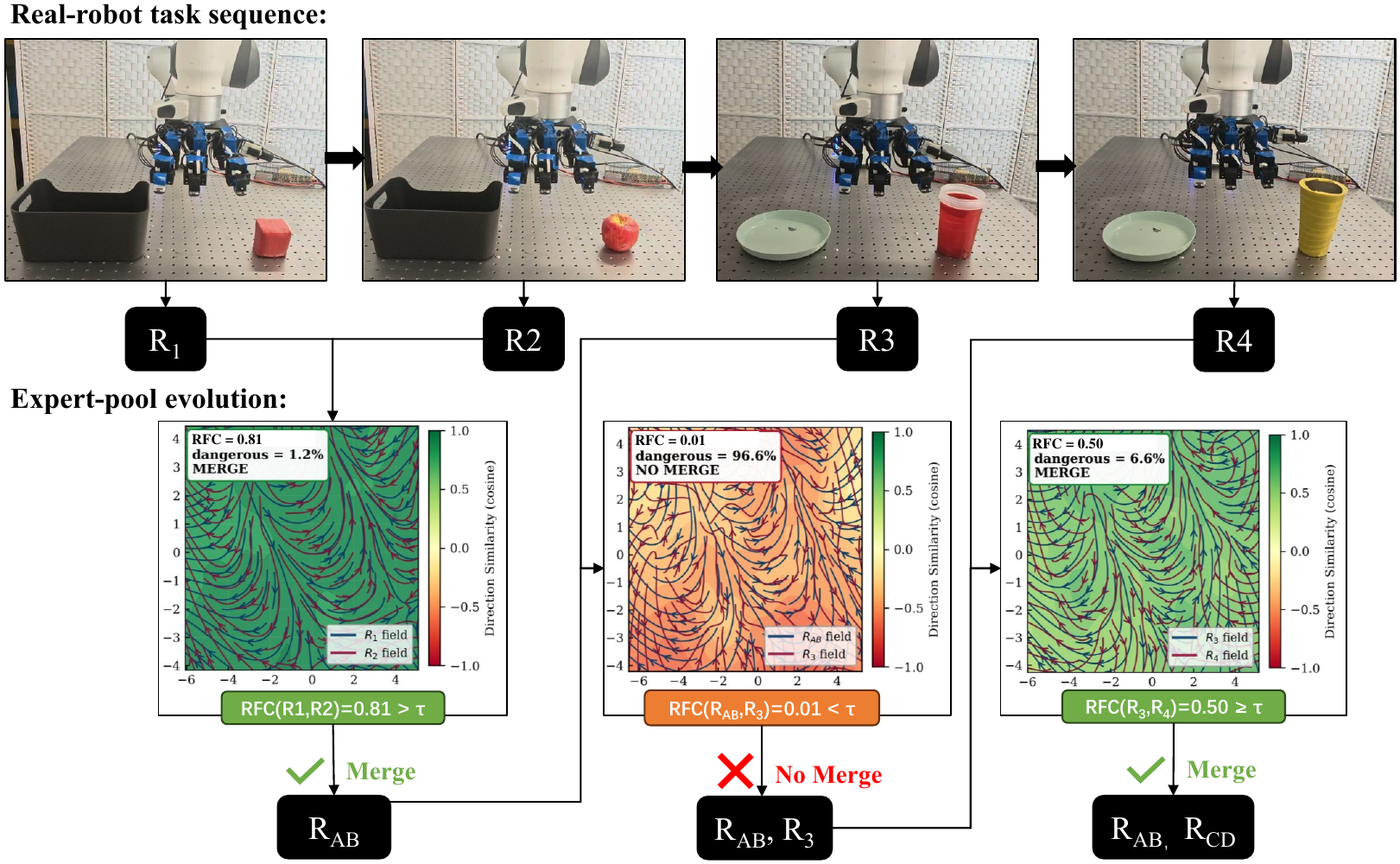}
    \caption{\textbf{Sequential RFC analysis on the real robot.} Streamlines show the compared correction fields; background color indicates local directional agreement (green) or opposition (red). Statistics use full observations and actions; the plotted projection is only a visualization. \textbf{(a)} RFC accepts $(R_1,R_2)$, producing $R_{AB}$. \textbf{(b)} RFC rejects $(R_{AB},R_3)$ and retains $R_3$. \textbf{(c)} RFC accepts $(R_3,R_4)$, producing $R_{CD}$. Two consolidated experts serve all four tasks.}
    \label{fig:rfc_realworld}
\end{figure*}

\subsection{Real-World Continual Learning}
\label{subsec:realworld}

\textbf{Task design.}
Fig.~\ref{fig:realtasks} shows the four tabletop tasks, arranged in two paired groups: $(T_1)$~\emph{Pinch-Cube} and $(T_2)$~\emph{Pinch-Apple} use pinch grasps to place small objects into a box; $(T_3)$~\emph{Power-Steel} and $(T_4)$~\emph{Power-Plastic} use power grasps to place cups onto a plate. Within each group, tasks share the grasp primitive and target container but differ in appearance, geometry, and friction. This grouping helps interpret RFC decisions; shared grasp primitives do not establish that policy parameters can be merged successfully.

\textbf{Real-robot implementation.}
We use a Franka Emika Panda, a 16-DoF Leap hand, and a front-mounted RealSense D435. The frozen IMLE BC base is trained on demonstrations from all four tasks. The residual consumes frozen ViT-S RGB features and proprioception, without object poses or other privileged simulator state.

The residual uses rank 16 and hidden dimensions $512,512,256$. Real-robot training uses separate actor and critic encoders, an actor learning rate of $10^{-5}$, critic and temperature learning rates of $3\times10^{-4}$, a five-critic ensemble, an LLM-based success classifier, and per-step action clipping. 

\textbf{Evaluation protocol.}
We use an $N{\times}N$ continual learning evaluation matrix with $N{=}4$. After training the residual for task $T_i$ and applying any RFC-gated merge, we evaluate the current residual set on all four tasks for 20 trials each, yielding entries $a_{i,j}$: diagonal entries use the just-trained (possibly merged) residual, lower-triangular entries use the residual associated with task $j$ (e.g., the merged $R_{AB}$), and upper-triangular entries report zero-shot success of the latest residual on as-yet-unseen tasks. The full $4{\times}4$ matrix appears in Appendix Table~\ref{tab:real_matrix}.

\textbf{Main results.}
Table~\ref{tab:realcl}(a) reports the three transfer metrics. With the IMLE+Leap-hand setup, ResMerge attains FWT~50\%, NBT~7 pp, and AUC~75\%, outperforming DMPEL on both stability (NBT) and overall performance (AUC). RFC selectively merges the residuals for $(T_1,T_2)$ into $R_{AB}$ and those for $(T_3,T_4)$ into $R_{CD}$, so the final residual pool contains two consolidated experts rather than four task-specific ones. Each consolidated expert serves the two tasks that share a grasp primitive.

\begin{table}[!t]
\caption{\textbf{Real-world evaluation.} We report continual learning metrics in (a) and average success with residual-expert counts in (b).}
\label{tab:realcl}

\centering
\renewcommand{\arraystretch}{1.0}
\footnotesize

\begin{tabular*}{\columnwidth}{
    @{}l|@{\extracolsep{\fill}}ccc@{}
}
\toprule
\multicolumn{4}{l}{\textbf{(a) Continual learning metrics}} \\
\textbf{Method}
& \textbf{FWT (\%)}
& \textbf{NBT (pp)}
& \textbf{AUC (\%)} \\
\midrule

DMPEL
& 24 & 23 & 54 \\

\resmergefullrow{blue!8}
\textbf{ResMerge}
& \textcolor{blue!70!black}{\textbf{50}}
& \textcolor{blue!70!black}{\textbf{7}}
& \textcolor{blue!70!black}{\textbf{75}} \\

\midrule
\multicolumn{4}{l}{\textbf{(b) Average success and residual-expert count}} \\
\textbf{Method}
& \multicolumn{2}{c}{\textbf{Avg. SR (\%)}}
& \textbf{Experts} \\
\midrule

TAIL
& \multicolumn{2}{c}{52.5}
& -- \\

\resmergefullrow{blue!8}
ResMerge (no merge)
& \multicolumn{2}{c}{
    \textcolor{blue!70!black}{\textbf{78.75}}
}
& 4 \\

\resmergefullrow{blue!8}
\textbf{ResMerge}
& \multicolumn{2}{c}{72.5}
& \textcolor{blue!70!black}{\textbf{2}} \\

\bottomrule
\end{tabular*}
\end{table}

\textbf{Additional real-robot baselines.}
TAIL~\citep{liu2024tailtaskspecificadaptersimitation} learns task-specific parameter-efficient adapters via imitation learning; no-merge keeps each learned residual as a separate expert. Table~\ref{tab:realcl}(b) reports average SR, which is distinct from AUC; a dash denotes an unreported expert count. In average SR, ResMerge reaches 72.5\%, compared with 52.5\% for TAIL and 78.75\% without merging. Consolidation retains 92.1\% of the no-merge average SR while halving the residual-expert count. The 6.25-percentage-point reduction in success quantifies the cost of this storage saving.

\textbf{Sequential RFC walk-through.}
Following the task setup in Fig.~\ref{fig:realtasks} and the results in Table~\ref{tab:realcl}, Fig.~\ref{fig:rfc_realworld} traces the sequential merge-or-expand decisions. \textbf{(a)} After learning $R_1$ and $R_2$, their fields show high consistency (RFC $0.81$, $1.25\%$ dangerous states), and RFC accepts the merge into $R_{AB}$. \textbf{(b)} After learning $R_3$, $R_{AB}$ and $R_3$ show low residual-field consistency (RFC $0.01$, $96.6\%$ dangerous states), so RFC rejects the merge and adds $R_3$ as a new expert. \textbf{(c)} After learning $R_4$, the fields show greater consistency (RFC $0.50$, $6.6\%$ dangerous states), and RFC accepts the merge into $R_{CD}$. The resulting pool $\{R_{AB},R_{CD}\}$ separates the pinch-grasp and power-grasp task groups, consistent with their shared corrective behavior.

\textbf{Interpreting the field visualization.}
RFC is evaluated on $2{,}000$ replay observations sampled with a fixed seed and shared by both compared experts. On the real robot it uses the full 836-dimensional observation representation and 22-dimensional residual actions. PCA is used only to display the fields in Fig.~\ref{fig:rfc_realworld}, not to compute RFC. The projection illustrates local directions; RFC uses direction and magnitude in the original action space. Agreement between residuals alone does not determine task success.

\textbf{Offline merge-all diagnostic.}
We evaluate forced merging offline only and do not report a closed-loop merge-all success rate. Cross-group pairs have mean DC 0.04--0.15 and 87--99\% conflicting states. Forced merging reduces action-direction fidelity from 0.916 to 0.693 on $T_1$ and from 0.792 to 0.661 on $T_2$. These diagnostics reveal interference in correction fields, rather than closed-loop task success.

\subsection{Ablation Results and Analysis}
\label{subsec:ablations}
We ablate merging and residual architecture on LIBERO. Learn-SR measures success after residual learning; Merge-SR measures success after merging and RL refinement.

\textbf{Merging Strategy.} On LIBERO-Object (ViT-T), Table~\ref{tab:ablation_strategy} compares \emph{no-merge}, \emph{merge-all}, and \emph{RFC-gated} (ours): no-merge retains separate experts, merge-all yields lower post-refinement SR, and RFC-gated achieves the highest post-refinement SR and FWT and the lowest NBT. These results evaluate the complete merging-and-refinement procedure.

\begin{table}[!htbp]
\caption{\textbf{Effect of merging strategy.}
Learn-SR and Merge-SR are mean success rates across all ten LIBERO-Object tasks with ViT-T.
Merge-SR includes the RL refinement that follows consolidation.}
\label{tab:ablation_strategy}

\centering
\renewcommand{\arraystretch}{1.05}

{\footnotesize
\begin{tabular*}{\columnwidth}{
    @{}l@{\extracolsep{\fill}}ccccc@{}
}
\toprule
\textbf{Strategy}
& \textbf{Learn-SR}
& \textbf{Merge-SR}
& \textbf{FWT}
& \textbf{NBT}
& \textbf{AUC} \\
\midrule

No Merge
& 36 & -- & 27 & 13 & 54 \\

Merge-All
& 36 & 21 & 23 & 22 & 37 \\

\resmergefullrow{blue!8}
\textbf{ResMerge}
& 36
& \textcolor{blue!70!black}{$\mathbf{78}$}
& \textcolor{blue!70!black}{$\mathbf{49}$}
& \textcolor{blue!70!black}{$\mathbf{-8}$}
& \textcolor{blue!70!black}{$\mathbf{79}$} \\

\bottomrule
\end{tabular*}
}
\end{table}

\begin{table}[!htbp]
\caption{\textbf{Effect of residual architecture.}
LoRA and MLP residuals are compared before and after merging and refinement.}
\label{tab:ablation_reisual_actor}

\centering
\renewcommand{\arraystretch}{1.05}

{\footnotesize
\begin{tabular*}{\columnwidth}{
    @{}l@{\extracolsep{\fill}}cc@{}
}
\toprule
\textbf{Residual Architecture}
& \textbf{Learn-SR (\%)}
& \textbf{Merge-SR (\%)} \\
\midrule

MLP Residual Policy
& 36
& 14 \\

\resmergefullrow{blue!8}
\textbf{LoRA Residual Policy (ours)}
& 36
& \textcolor{blue!70!black}{$\mathbf{78}$} \\

\bottomrule
\end{tabular*}
}
\end{table}

\textbf{Residual architecture: MLP vs.~LoRA.} Both architectures achieve 36\% Learn-SR (Table~\ref{tab:ablation_reisual_actor}), but their post-merge SR differs: the MLP residual achieves 14\%, while the LoRA residual achieves 78\%. These results favor low-rank residual actors when continual adaptation includes merging and refinement across successive tasks.

Additional comparisons of RFC thresholds and merging metrics appear in the Appendix (Tables~\ref{tab:rfc_thresholds} and~\ref{tab:ablation_metric}).

\section{Conclusion}
We present ResMerge, which combines residual RL without additional expert demonstrations during adaptation with RFC-guided merging of low-rank residual policies. A consolidated expert can serve multiple tasks, while incompatible candidates remain separate. Experiments on LIBERO across two backbones and on a real-world Franka robot show improved continual learning performance over strong PECFT baselines. RFC shifts compatibility assessment from task space to residual action-field space, enabling selective consolidation to reduce stored residual parameters.

\section{Limitations}
ResMerge assumes useful initial base-policy competence and studies refinement of existing skills. Adaptation requires no additional expert demonstrations, but relies on online interaction, resets, reward signals, and post-merge refinement on affected tasks. Developing more sample-efficient residual RL for continual adaptation remains an open direction. Adaptive RFC threshold selection also remains future work beyond the fixed per-backbone thresholds evaluated here.

\appendix[Additional Experimental Results]
\label{appendix:additional}
\textbf{Implementation settings.}
OpenVLA-OFT simulation and the real-robot setup use different learning rates. Table~\ref{tab:implementation_settings} summarizes these settings. The real-robot residual uses rank 16, hidden dimensions $512,512,256$, and a five-critic ensemble. Both simulation and the real robot use residual scale $\alpha=0.05$. We report these settings separately to distinguish the two platforms.

\begin{table}[!htbp]
\caption{\textbf{Implementation settings by experimental platform.} Real-robot architecture and evaluation details are specified separately in the text.}
\label{tab:implementation_settings}
\centering
{\footnotesize
\begin{tabular*}{\columnwidth}{@{}l@{\extracolsep{\fill}}cc@{}}
\toprule
\textbf{Parameter} & \textbf{OpenVLA-OFT} & \textbf{Real robot} \\
\midrule
Actor learning rate & $3\times10^{-4}$ & $10^{-5}$ \\
Critic learning rate & $3\times10^{-4}$ & $3\times10^{-4}$ \\
Residual scaling factor $\alpha$ & 0.05 & 0.05 \\
\bottomrule
\end{tabular*}}
\end{table}

\textbf{RFC threshold sensitivity.}
Table~\ref{tab:rfc_thresholds} studies RFC thresholds on LIBERO-Object (ViT-T). At $\tau=0.1$, the metrics match merge-all; at $\tau=0.7$, FWT is 27\% and AUC is 47\%. Both intermediate thresholds retain FWT of 49\%, while increasing $\tau$ from 0.3 to 0.5 improves NBT from $-2$ to $-8$ pp and AUC from 71\% to 79\%. These results favor selective consolidation at an intermediate threshold.

\begin{table}[!htbp]
\caption{\textbf{RFC threshold sensitivity.}
We vary $\tau$ on LIBERO-Object with ViT-T, keeping it fixed throughout each task sequence.}
\label{tab:rfc_thresholds}

\centering

{\footnotesize
\begin{tabular*}{\columnwidth}{
    @{\extracolsep{\fill}}c|ccc@{}
}
\toprule
\textbf{Threshold $\tau$}
& \textbf{FWT}
& \textbf{NBT}
& \textbf{AUC} \\
\midrule

0.1 & 23 & 22 & 37 \\
0.3 & \textbf{49} & $-2$ & 71 \\
0.5 & \textbf{49} & $\mathbf{-8}$ & \textbf{79} \\
0.7 & 27 & 22 & 47 \\

\bottomrule
\end{tabular*}
}
\end{table}

\textbf{Merging metric.} Table~\ref{tab:ablation_metric} compares task similarity with RFC on LIBERO-Object (ViT-T). The task-similarity baseline uses a GPT vision-language model to judge task similarity from language descriptions and several visual frames, and merges residual experts when the model judges their tasks to be similar. Both begin at 36.0\% pre-merge success. RFC reaches 78.0\% after merging and refinement, compared with 40.0\% for task similarity, and improves transfer and retention. This comparison supports assessing compatibility between residual corrections rather than between tasks alone.

\begin{table}[!htbp]
\caption{\textbf{Comparison of merging metrics.} Results are on LIBERO-Object with ViT-T. Post-merge success includes RL refinement.}
\label{tab:ablation_metric}
\centering
{\footnotesize
\begin{tabular*}{\columnwidth}{@{}l@{\extracolsep{\fill}}ccccc@{}}
\toprule
\textbf{Metric} & \textbf{Pre-SR} & \textbf{Post-SR} & \textbf{FWT} & \textbf{NBT} & \textbf{AUC} \\
\midrule
Task Similarity & 36.0 & 40.0 & 26.0 & 11.0 & 46.0 \\
\resmergefullrow{blue!8}
\textbf{RFC} & 36.0 & \textbf{78.0} & \textbf{49.0} & $\mathbf{-8.0}$ & \textbf{78.7} \\
\bottomrule
\end{tabular*}}
\end{table}

\textbf{Real-robot evaluation matrix.} Table~\ref{tab:real_matrix} reports success after each stage on all four tasks. Diagonal entries evaluate the just-trained skill, lower-triangular entries measure retention, and upper-triangular entries measure transfer to unseen tasks. The first two skills share $R_{AB}$ after stage 2; the last two share $R_{CD}$ after stage 4. The final pool thus retains four skills with two residual experts.

\begin{table}[!htbp]
\caption{\textbf{Real-robot success across the task sequence.} Rows indicate the completed training stage and residual pool; columns indicate the evaluated task. Each cell is evaluated over 20 trials. Success rates are percentages.}
\label{tab:real_matrix}
\centering
{\footnotesize
\begin{tabular*}{\columnwidth}{@{}l@{\extracolsep{\fill}}cccc@{}}
\toprule
\textbf{Stage and residual pool} & $\boldsymbol{T_1}$ & $\boldsymbol{T_2}$ & $\boldsymbol{T_3}$ & $\boldsymbol{T_4}$ \\
\midrule
$T_1$: $\{R_1\}$ & \textbf{85} & 35 & 55 & 55 \\
$T_2$: $\{R_{AB}\}$ & 75 & \textbf{60} & 55 & 50 \\
$T_3$: $\{R_{AB},R_3\}$ & 75 & 60 & \textbf{85} & 50 \\
$T_4$: $\{R_{AB},R_{CD}\}$ & 75 & 60 & 75 & \textbf{80} \\
\bottomrule
\end{tabular*}}
\end{table}

\bibliographystyle{IEEEtranN}
\bibliography{references/references}

\end{document}